\documentclass[letterpaper, 10 pt, conference]{ieeeconf}
\IEEEoverridecommandlockouts                              
\PassOptionsToPackage{hyphens}{url}\usepackage{hyperref}
\usepackage{lettrine}
\usepackage{graphicx}
\usepackage{float}
\usepackage[margin=0.75in]{geometry}
\usepackage{tikz}
\usepackage[caption=false]{subfig}
\usepackage{amsmath}
\usepackage{amssymb}
\usepackage{bbm}
\usepackage{tabularx}
\usepackage{dsfont}
\DeclareMathOperator*{\argmax}{arg\,max}

\title{Evaluating Uncertainty Quantification in End-to-End Autonomous Driving Control}
\author{\authorblockN{Rhiannon Michelmore\authorrefmark{1}, Marta Kwiatkowska\authorrefmark{1}, Yarin Gal\authorrefmark{1}}
        \authorblockA{\authorrefmark{1}University of Oxford, UK}
        \authorblockA{\authorrefmark{1}\texttt{firstname.lastname@cs.ox.ac.uk}}
}

\begin{document}
\maketitle
\begin{abstract}
	A rise in popularity of Deep Neural Networks (DNNs), attributed to more powerful GPUs and widely available datasets, has seen them being increasingly used within safety-critical domains. One such domain, self-driving, has benefited from significant performance improvements, with millions of miles having been driven with no human intervention. Despite this, crashes and erroneous behaviours still occur, in part due to the complexity of verifying the correctness of DNNs and a lack of safety guarantees.

    In this paper, we demonstrate how quantitative measures of uncertainty can be extracted in real-time, and their quality evaluated in end-to-end controllers for self-driving cars. To this end we utilise a recent method for gathering approximate uncertainty information from DNNs without changing the network's architecture. We propose evaluation techniques for the uncertainty on two separate architectures which use the uncertainty to predict crashes up to five seconds in advance. We find that mutual information, a measure of uncertainty in classification networks, is a promising indicator of forthcoming crashes.

\end{abstract}

\section{Introduction}

Deep learning, and in particular Deep Neural Networks (DNNs), have seen a surge in popularity over the past decade, and their use has become widespread in many fields. This increase in popularity, attributed to (i) more powerful GPU implementations and (ii) the availability of large amounts of data, has led to significant performance gains. 
DNNs are now being deployed
in safety-critical systems such as medical diagnosis and, in particular, autonomous cars. The latter 
are computationally efficient and have driven millions of miles without human intervention \cite{Waymo,CADis}, 
but offer few safety guarantees. Our lack of understanding of how DNNs work \cite{zeiler2014visualizing}, paired with the prohibitive difficulty of verifying the correctness of DNNs of this magnitude \cite{huang2017safety}, has led to erroneous edge-case behaviours and unforeseen consequences. Most notably, there have been crashes involving autonomous cars that were a direct result of the self-driving system malfunctioning. In May 2016, the autopilot feature of a Tesla Model S caused a fatal accident when it failed to distinguish the white side of a truck against the bright sky \cite{TeslaCrash}. It is clear that, although 
DNNs are said to perform 
as well as humans, there are still erroneous edge-case behaviours which need to be detected, analysed and ultimately eliminated.

In this paper, we focus on end-to-end controllers for self-driving cars, that is, DNNs mapping raw pixels from a front-facing camera directly to steering instructions. End-to-end learning-based approaches have been used in several existing autonomous vehicle control systems, including 
the DARPA Autonomous Vehicle (DAVE) project \cite{lecun2004dave}, and more recently in NVIDIA's PilotNet (formally known as DAVE-2) \cite{bojarski2016end}. The motivation for using an end-to-end controller is to remove the need for complex, specifically coded scenarios, instead allowing the network to define its own features based on 
training.

When the entire control system is an end-to-end controller (a DNN), as opposed to a collection of subsystems, techniques such as uncertainty estimation can be employed to more accurately assess the controllers confidence in the decisions \cite{mcallister2017concrete}. 
\emph{Model uncertainty} is a measure of how unsure a DNN model is in its prediction, and can be used to understand if a model is under- or over-confident, as well as to determine regions of input where more training data is required \cite{Gal2016Uncertainty}. With certain DNN activation functions such as ReLU, model uncertainty increases as the input moves further from the training data; this information can be used to augment the training data accordingly. A recent technique from Gal and Ghahramani \cite{gal2016dropout} provides a simple, \emph{real-time} method to extract an estimation of model uncertainty using any stochastic regularisation technique, which are a common feature of modern DNN models. This technique, and dropout, our stochastic regularisation technique of choice, will be explained in detail in Section~\ref{subsec:uncert}. The motivation behind modelling uncertainty is to improve \textit{safety} by creating systems that take into account the confidence of the model at each stage to avoid error propagation \cite{mcallister2017concrete}. A meaningful measure of system uncertainty can be used as a basis for safe decisions.

This paper proposes quantitative evaluations of uncertainty for use 
within end-to-end controllers for self-driving cars. The key contributions are:
\begin{itemize}
\item We demonstrate how quantitative measures of uncertainty can be extracted in real-time from end-to-end controllers. 
    \item We show how uncertainty thresholds can be chosen and used to alert the operator to areas of low model confidence.
    \item We evaluate the techniques on 
    two modified PilotNet \cite{bojarski2017explaining} architectures, for both regression and classification settings, within a driving simulator. 
    We demonstrate how to train these networks and how to select hyper-parameters.
	\item We present preliminary results that show significant changes in uncertainty, specifically mutual information, up to five seconds before crashes.
\end{itemize}

\section{Related Work}
With the exception of \cite{kendall2015bayesian,kendall2016modelling}, DNN-based approaches for autonomous driving do not often consider model uncertainty. 
Recent work by Yang et al \cite{yang2018end} has seen the addition of discrete speed control prediction, along with steering angle prediction, to end-to-end controllers for self-driving cars. Their work aims to make DNN-based controllers more viable, as steering angle alone is not sufficient for vehicle control. The resulting multi-modal multi-task network was shown to predict both steering angle and speed commands accurately, but does not include the use of uncertainty for any means.

Kendall et al's paper on pixel-wise semantic segmentation \cite{kendall2015bayesian} utilised model uncertainty to improve segmentation performance. In addition to this, they were able to show that the highest areas of uncertainty occurred on class boundaries. These results were reinforced by Kampffmeyer et al \cite{kampffmeyer2016semantic} which considers several other methods and also concludes that uncertainty maps are a good measure of uncertainty in segmented images.

In 2016, Kendall and Cipolla \cite{kendall2016modelling} developed tools for the localisation of a car given a forward facing photo. They found that model uncertainty correlated to positional error; test photos with strong occlusion resulted in high uncertainty and the uncertainty displayed a linearly increasing trend with the distance from the training set.

\section{Background}
\subsection{End-to-end controllers for self-driving}
An end-to-end controller is a controller in which the end-to-end process, from sensors to actuation, involves a single DNN without modularisation. In the context of self-driving, the sensors might include camera input, infrared (IR) sensors, light detection and range sensors (LiDAR), or a combination of these in addition to many others. The outputs are typically steering angle, braking and acceleration values. In this paper, we focus on up to three camera inputs, placed on the front of the car facing forwards. The input to the network is therefore up to three images, and the output is the desired steering angle.

A typical feed-forward DNN consists of \textit{layers} of \textit{neurons}; these neurons are connected via edges to neurons in different layers. Each edge has a corresponding weight and each neuron sums together the product of each input $x$ and edge weight $W$, then applies a non-linear activation function $f$ over the result: $output = f(\mathbf{xW})$. Common choices for activation functions include the sigmoid function \cite{cybenko1989approximation} and the Rectified Linear Unit (ReLU) \cite{nair2010rectified}. Networks intended for regression tasks have one output per continuous value to be predicted, and a common loss function for optimisation is the mean squared loss. Classification problems, in general, have as many neurons in the output layer as classes, and the final layer's activation function is a ``softmax'' function (see Equation~\ref{equation:softmax}). The most common loss function for classification is the cross-entropy loss. For more detail, we refer the interested reader to \cite{goodfellow2016deep}.

\begin{equation}
\text{softmax}(\mathbf{z}) := [\frac{e^{z_1}}{\sum_{k=1}^Ne^{z_k}},...,\frac{e^{z_N}}{\sum_{k=1}^Ne^{z_k}}]
\label{equation:softmax}
\end{equation}

Convolutional neural networks (CNNs) are commonly used in self-driving and other image recognition tasks as they reduce the number of network parameters, reduce training time and prevent overfitting \cite{goodfellow2016deep}. CNNs differ from DNNs in that they include \textit{convolution layer(s)}. The output of a neuron in a convolution layer is computed using only a small region (\textit{window}) of the layer before it is combined using a convolution kernel. This closely resembles the human visual system, suggesting that CNNs are well suited for vision based tasks \cite{goodfellow2016deep}.

\subsection{Bayesian uncertainty estimation}\label{subsec:uncert}

In many fields, uncertainty is used to determine the dependability of a model. In self-driving, if the training set of a model consisted of only images from highways and the model was then given an image of a dirt track, the model would return a steering angle but we would ideally require the model to have high \textit{uncertainty} as it would not have seen this type of image before. In classification problems, the softmax probabilities are not enough to indicate whether the model is confident in its prediction, as a standard model would pass the predictive mean (a point estimate) through the softmax rather than the whole predictive distribution \cite{deepmodelblog}. This leads to high probabilities (confidence) on points far from the training data.

The above was an example of out-of-distribution test data. Other examples of sources of uncertainty include noisy data, and situations where many models explain the same dataset equally (model parameter uncertainty). Noisy data is an example of aleatoric uncertainty, whereas model parameter uncertainty is an example of model uncertainty (or epistemic uncertainty) \cite{Gal2016Uncertainty}, the confidence the model has in its prediction, which is what this paper focuses on.

A recent technique from Gal and Ghahramani \cite{gal2016dropout} allows the gathering of approximate uncertainty information from DNNs without changing the architecture (given that some stochastic regularisation technique such as dropout has been used). Dropout is a regularisation technique that sets 1-\textit{p} proportion of the dropout layers' input to be 0, where \textit{p} is the dropout probability \cite{hinton2012improving}. The dropped weights are often scaled by 1/\textit{p} to maintain constant output magnitude.

It has been shown that a network that uses dropout is an approximation to a Gaussian process \cite{gal2016dropout}. Equation~\ref{equation:predictgauss} shows how a prediction can be performed with a Gaussian process, where \textbf{f} is the space of functions, \textbf{X} is the training data and \textbf{Y} is the training outputs. The expectation of \textbf{y*} is called the predictive mean of the model and the variance is the predictive uncertainty.

\begin{equation}
p(\mathbf{y^*}|\mathbf{x^*,X,Y}) = \int p(\mathbf{y^*}|\mathbf{f^*})p(\mathbf{f^*}|\mathbf{x^*,X,Y})\mathrm{d}\mathbf{f^*}
\label{equation:predictgauss}
\end{equation}

For a regression network, Equations \ref{equation:mean} and \ref{equation:var} are used to obtain an approximation of the predictive mean and variance of the Gaussian process that the network is an approximation of.

\begin{equation}
\mathbb{E}(\mathbf{y^*}) \approx \frac{1}{T} \sum_{t=1}^{T}\widehat{\mathbf{y}}_t^*(\mathbf{x^*})
\label{equation:mean}
\end{equation}

\begin{equation}
\begin{split}
\text{Var}(\mathbf{y^*}) \approx \tau^{-1}\mathbf{I}_D &+ \frac{1}{T}\sum_{t=1}^{T}\widehat{\mathbf{y}}_t^*(\mathbf{x^*})^T\widehat{\mathbf{y}}_t^*(\mathbf{x^*}) \\
&- \mathbb{E}(\mathbf{y^*})^T\mathbb{E}(\mathbf{y^*})
\end{split}
\label{equation:var}
\end{equation}

The set $\{\widehat{\mathbf{y}}_t^*(\mathbf{x^*})\}$ of size $T$ is the results from $T$ stochastic forward passes through the network. It is important that the non-determinism from dropout is retained at prediction time to ensure different units will be dropped per pass through. Relating back to the Gaussian process, these are empirical samples from the approximate predictive distribution seen in Equation~\ref{equation:predictgauss}. $\tau$ relates to the precision of the Gaussian process model, and is used in the calculation of the predictive variance. $\tau$ can be calculated as seen in Equation~\ref{equation:tau}, where $l$ is a user-defined length scale, $p$ is the probability of units \textit{not} being dropped, $N$ is the number of training samples and $\lambda$ is  multiplier used in the L\textsubscript{2} regularisation of the network.

\begin{equation}
\tau = \frac{l^2p}{2N\lambda}
\label{equation:tau}
\end{equation}

A small length-scale (corresponding to high frequency data) with high $\tau$ (corresponding to small observation noise) will lead to a small weight-decay, which might mean the model fits the data well but generalises badly. Conversely, a large length-scale and low $\tau$ will lead to strong regularisation. There is a trade-off between length-scale and model precision. In practice, the model precision $\tau$ is often found by grid searching over the weight decay $\lambda$ to minimise validation error, choosing a length-scale that correctly describes the data, and then putting the values into Equation~\ref{equation:tau}. It can also be found by grid searching over $\tau$ values directly. 

For classification tasks, there are several methods of obtaining uncertainty information. As previously mentioned, softmax probabilities are a poor indicator as they are the result of a single \textit{deterministic} pass of a point estimate through the network, which can lead to high confidence on points far from the training data \cite{Gal2016Uncertainty}. The three approaches used in this paper to summarise classification uncertainty are \textit{variation ratios} \cite{freeman1965elementary}, \textit{predictive entropy} \cite{shannon2001mathematical} and \textit{mutual information} \cite{shannon2001mathematical}.

\textit{Variation ratio} is a measure of dispersion, its value is high when classes are more equally likely and low when there is a clear winner. Variation ratio, as with predictive uncertainty in regression tasks, requires $T$ stochastic forward passes through the network for a test input $\mathbf{x}$. A set of $T$ labels $y_t$ is collected, where $y_t$ is the class with the highest softmax output of that pass through. The mode of the distribution $c^*$ and the number of times it was sampled $f_x$ can then be used to obtain the variation ratio. These calculations can be seen in Equations \ref{equation:mode}, \ref{equation:freq} and \ref{equation:varrat}.

\begin{equation}
c^* = \argmax_{c=1,...,C}\sum_t\mathds{1}[y^t=c]
\label{equation:mode}
\end{equation}

\begin{equation}
f_x = \sum_t\mathds{1}[y^t=c^*]
\label{equation:freq}
\end{equation}

\begin{equation}
\text{variation-ratio}[x] := 1 - \frac{f_x}{T}
\label{equation:varrat}
\end{equation}

\textit{Predictive entropy} captures the average amount of information present in the predictive distribution, $\mathbb{H}[y|\mathbf{x},\mathcal{D}_{\text{train}}]$. In our setting the predictive entropy can be approximated by collecting the softmax probability vectors over $T$ stochastic forward passes, and for each class, averaging the softmax probability and multiplying it by the log of that average. This can be seen in Equations~\ref{equation:smax} and \ref{equation:entropy}, where $\mathbf{f^{\omega}}$ is the network with model parameters $\omega$.

\begin{equation}
\begin{split}
&\text{softmax}(\mathbf{f^{\omega}(x)}) := \\ & \quad[p(y=1|\mathbf{x,\widehat{\omega}_t}),...,p(y=C|\mathbf{x,\widehat{\omega}_t})] \\
\end{split}
\label{equation:smax}
\end{equation}
\begin{equation}
\begin{split}
\widetilde{\mathbb{H}}[y|\mathbf{x},\mathcal{D}_{\text{train}}] := &-\sum_c(\frac{1}{T}\sum_t p(y=c|\mathbf{x,\widehat{\omega}_t}))\\
&\cdot \log(\frac{1}{T}\sum_t p(y=c|\mathbf{x,\widehat{\omega}_t}))
\end{split}
\label{equation:entropy}
\end{equation}

The final measure is \textit{mutual information}, $\mathbb{I}[y,\omega|\mathbf{x},\mathcal{D}_{\text{train}}]$. Test points that maximise mutual information are points on which the model is uncertain on average, but there are model parameters that erroneously produce high confidence predictions. Mutual information is calculated similarly to predictive entropy, but with an extra term as seen in Equation~\ref{equation:mutinf}.

\begin{equation}
\begin{split}
\widetilde{\mathbb{I}}[y,\omega|\mathbf{x},\mathcal{D}_{\text{train}}] := (&-\sum_c(\frac{1}{T}\sum_t p(y=c|\mathbf{x,\widehat{\omega}_t}))\\
&\cdot log(\frac{1}{T}\sum_t p(y=c|\mathbf{x,\widehat{\omega}_t})))\\
&+ \frac{1}{T}\sum_{c,t}p(y=c|\mathbf{x,\widehat{\omega}_t})log p(y=c|\mathbf{x,\widehat{\omega}_t})
\end{split}
\label{equation:mutinf}
\end{equation}

Variation ratios and predictive entropy are both measures of predictive uncertainty, whereas mutual information is a measure of the model's confidence in its output. Further information on this can be found in \cite{Gal2016Uncertainty}. Having multiple measures of uncertainty is arguably 
more powerful than the sole measure available for regression tasks, as it allows for different types of uncertainty to be captured and gives us more information about the performance of the model.

\section{Methodology} 
To investigate evaluation techniques of uncertainty in end-to-end controllers for self-driving, we must (i) explore the different types of uncertainty available from different network architectures, (ii) assess the suitability of each network architecture as a function of accuracy and uncertainty type, (iii) calibrate thresholds for these uncertainties, and (iv) study uncertainty levels in real-time in a simulator.

\subsection{Network setup}
As mentioned above, the inputs in our scenario are images, of size 66x200x3 (RGB colour channels are retained), from the Udacity self-driving car simulator \cite{udacitysim}. Our desired output is a steering angle. The simulator only allows for angles between -25 and +25 degrees so the network is limited to this range of values. We use two networks in this work, one that treats the problem of predicting a steering angle as a regression problem and one that treats it as a classification problem.

Traditionally, steering angle prediction has been treated as a regression problem. However, it has been shown that posing regression tasks as classification tasks often shows improvement over direct regression training \cite{rothe2015dex}. In addition to this, although theoretically continuous, steering angle in the real-world is commonly a discrete variable due to mechanical limitations. 

\begin{figure}
\includegraphics[width=0.5\textwidth]{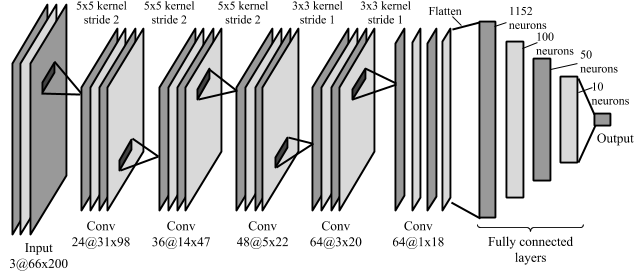}
\caption{The architecture of our regression network. ReLU is used as the activation function, and dropout is applied after every layer but the first and last with probability $p=0.05$.}
\label{figure:regressionarch}
\end{figure}

Both architectures are heavily based on NVIDIA's end-to-end self-driving controller PilotNet \cite{bojarski2017explaining}. The architecture of the regression network, seen in Figure~\ref{figure:regressionarch}, has additional dropout layers after every layer but the first and last, with probability $p=0.05$. An L\textsubscript{2} regularizer with scale factor $\lambda = 1e-6$ was used on every layer but the final, and the ReLU activation function ($relu(x) = max(0,x)$) was used throughout. The loss function used was mean square error.

In calculating $\tau$,  the value $l = 0.01$ was selected after a grid search experiment, because it, along with the value selected for $\lambda$, produced the highest accuracy out of the values searched over and matched values used in similar experiments \cite{gal2016dropout}.

\begin{figure}
\includegraphics[width=0.5\textwidth]{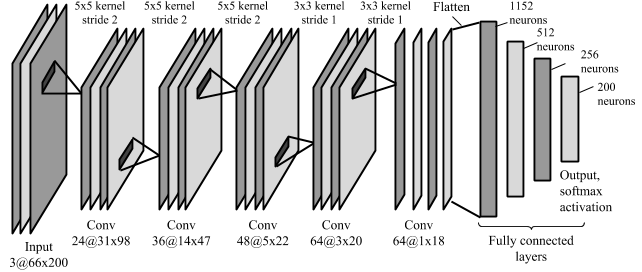}
\caption{The architecture of our classification network.}
\label{figure:classarch}
\end{figure}

In order to frame the problem of predicting steering angles as a classification problem, the training angles were bucketed into one of two-hundred intervals (classes). As the simulator has a limit of -25 degrees to +25 degrees, each bucket has a precision of 0.25 degrees. The architecture of the classification network can be seen in Figure~\ref{figure:classarch}; it also uses ReLU activation functions, but has a softmax final layer and uses the categorical cross-entropy loss function.

\subsection{Network training}
Both networks were trained on a dataset of 24,496 images taken from the front centre of the car, with a random 20\% reserved for testing. 8037 of these images came from the training dataset included in the Udacity self-driving car challenge \cite{udacitysim}, the rest from data collected by the authors. The data was then augmented by mirroring each image horizontally and multiplying the steering angles by -1. For the classification task, the steering angles were further bucketed into their closest 0.25 degree interval. The speed of the car in both data collection and testing was around 15mph.

Further data augmentation can be done, as the simulator produces images from cameras at the left and right of the car, as well as the front centre. These additional images can be used with the original steering angle +/- a small correction resulting from the deviation from the centre of the car to the camera. It was found that, in this case, adding the extra images did not change the accuracy of the network sufficiently enough to include them so they were omitted.

Both networks were trained for 50 epochs with batch size 128.

\subsection{Uncertainty extraction}
At test time, each image $x$ was copied $T$ times into an array $\{x_1,...,x_T\}$ which was passed to the network for prediction. Non-determinism was retained by running the network in training mode. Higher $T$ increases processing time but returns a more accurate representation of the predictive distribution. The value of $T$ used here was 128 to match a single batch size; this provides a good trade-off between processing time and accuracy.

For regression, this resulted in a 128 length vector where the returned prediction was the mean (as in Equation~\ref{equation:mean}) and the variance was calculated according to Equation~\ref{equation:var}. The value of $\tau$ was 0.00328.

For classification, this resulted in a 128x200 matrix where each of the 128 rows represents the softmax output for that particular pass through the network. The mode of the maximum value in each row was taken to be the steering angle prediction, and the three types of uncertainty defined in Section~\ref{subsec:uncert} were calculated as seen in Equations \ref{equation:varrat}, \ref{equation:entropy} and \ref{equation:mutinf}.

We found that extracting uncertainty information in real-time was achievable if the number of stochastic forward passes was limited to a single batch size. The simulator sends an average of 6 frames per second to the receiver and we were able to consistently match this rate on a desktop PC with an Intel Core i5-6600 processor and 16GB RAM.

\subsection{Uncertainty evaluations}
In order for uncertainty information to be useful, we examined two scenarios: whether any uncertainty measure is significantly higher in places where the predicted angle is visually incorrect, and whether any uncertainty measure is significantly higher before a crash. These two scenarios define our evaluation metrics. 

\subsubsection{Evaluation Metric One}
For the first of the two investigations, manually labelled ground truth data was collected by writing a program to overlay the networks' predicted angles on the input images. It was then possible to manually decide whether the angle was ``safe'' or would lead to a crash, which was recorded to disk.

\begin{figure}
\centering
\subfloat[]{
\includegraphics[width=0.24\textwidth]{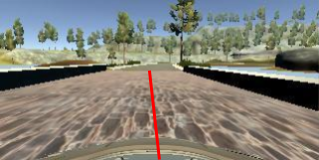}}
\subfloat[]{
\includegraphics[width=0.24\textwidth]{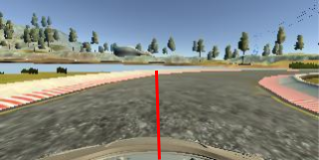}}
\caption{(a) An image generated by the manual labelling program and classification network that was marked ``safe''. (b) An image marked ``unsafe''.}
\label{figure:simline}
\end{figure}

The criterion for ``safe'' was if, at the end of a straight line from the centre of the car at the predicted angle, the car did not deviate from the road (see Figure~\ref{figure:simline}). The length of the straight line represents roughly 3 seconds of travel in the same direction but this will vary as the simulator car does not always travel at a constant speed and this cannot be controlled (only throttle value can be specified).

Another criterion for ``safe'' was briefly explored, where a curved line was drawn, to represent the car continuing to turn at the specified steering angle.

After manually labelling a set of 200 randomly selected images from the test set, ROC curves comparing the true and false positive rates for a range of uncertainty thresholds were generated. If the uncertainty for an ``unsafe'' image was above the threshold, it was marked as a true positive, whereas if the uncertainty was above the threshold for a ``safe'' image it was marked as a false positive.

\subsubsection{Evaluation Metric Two}
In the second scenario, evaluating whether uncertainty was significantly higher before a crash, involved running the simulator and recording uncertainty until a crash occurred. At that point, the uncertainty value from $n$ seconds before the crash was recorded, for $n=1,2,3,4,5,6$, and paired with a ``crashed'' label. Additional ``not crashed'' data was recorded, where uncertainty from $n$ seconds before a normal driving state was paired with a ``not crashed'' label. This data, for each $n$, was used to generate ROC curves to determine both the best threshold for uncertainty to predict ``crashed'' states that will occur in $n$ seconds, and to determine the most informative time $n$ before a crash.

\section{Results}
\subsection{Network performance}
The best version of the regression network achieved an RMSE of 0.1107 when using the predictive mean, a slight improvement over 0.1211 when predicting deterministically. In the simulator, the network drove around the track with an average of two crashes per loop, but its movement was jerky.

For classification, the best iteration of the network achieved an accuracy of 67\% (and did not change when using the mode of predictive distribution). This low accuracy could be explained by the fact that the steering angles needed to be converted to classes for classification, therefore losing some granularity. It is also worth noting that wrong predictions were frequently only a few classes away, so the predicted angle may still have been classified as ``safe''. Despite this accuracy, the car consistently drove around the simulator track with an average of zero to one crash per loop, and although still jerky, it was smoother than the regression network.

\subsection{Incorrect angle prediction}
\begin{figure}
\centering
\includegraphics[width=0.37\textwidth]{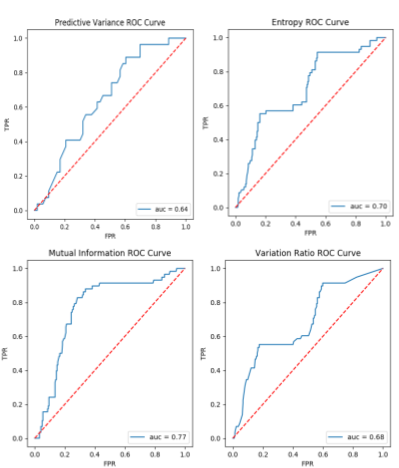}
\caption{The ROC curves for uncertainty in regression (top left) and classification (remaining plots) for incorrect angle prediction.}
\label{figure:classroc}
\end{figure}

The ROC curve for the regression network can be seen in the first graph in Figure~\ref{figure:classroc}. Using predictive variance to judge safe and unsafe road situations is only a slight improvement on random guessing (AUC = 0.64 versus 0.5). The ROC curves for each of the different uncertainty measures for classification can also be seen in Figure~\ref{figure:classroc}. It is clear that mutual information is most promising, with a high AUC value of 0.77.

The most important value to minimise in self-driving and crash prediction is false negatives (labelling unsafe situations as safe) as these will lead to crashes. With this in mind, a threshold with a high true positive rate was chosen to minimise this value. The threshold for mutual information was chosen as 0.612 which has a true positive rate of 0.81 and a false positive rate of 0.28. The threshold for entropy was chosen as 3.51 and the threshold for variation ratio was 0.75.

\subsection{Crash prediction}
\begin{figure}
\centering
\includegraphics[width=0.42\textwidth]{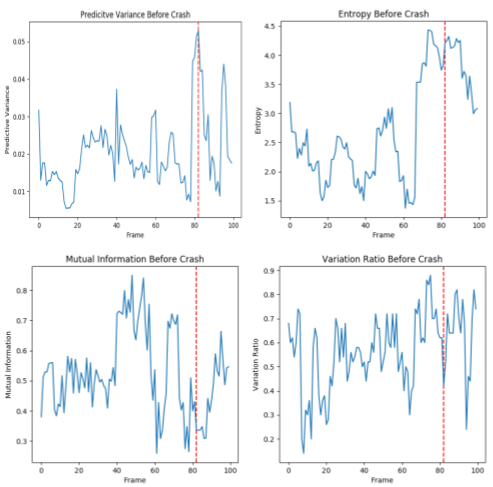}
\caption{The different uncertainty measures a number of frames before a crash occurs. The red dashed line indicates the frame at which the crash occurred.}
\label{figure:uncertbefore}
\end{figure}

The graphs in Figure~\ref{figure:uncertbefore} show the value of uncertainty for both regression and classification over the time period leading to one of the crashes; the red line indicates the frame at which the crash happened. Mutual information was once again the strongest indicator of incorrect behaviour, in this case meaning the car crashing. The mutual information in this graph peaks at around 27 frames, or 4.5 seconds, before the crash. Over the recorded crashes, the same uncertainty peak can be seen from between 7-31 frames, 1.17-5.17 seconds, before the crash. Table~\ref{table:timebeforecrash} shows for the first five crashes, the number of frames before the crash that the threshold was first passed, and the location of the most defined peak of mutual information before the crash.

Figure~\ref{figure:uncertbefore} shows the ROC curves for mutual information for 2, 3, 4 and 5 seconds before the ``crashed'' or ``not crashed'' event occurred, including values 0.25s either side (i.e. ``2 seconds before'' encompasses from 1.75s to 2.25s). Three seconds after a high mutual information value was recorded was the most likely time for a crash to occur, and the threshold for mutual information for this time step was set to be 0.501, which had a true positive rate of 73\% and a false positive rate of 28\%. 

\begin{figure}
\centering
\includegraphics[width=0.42\textwidth]{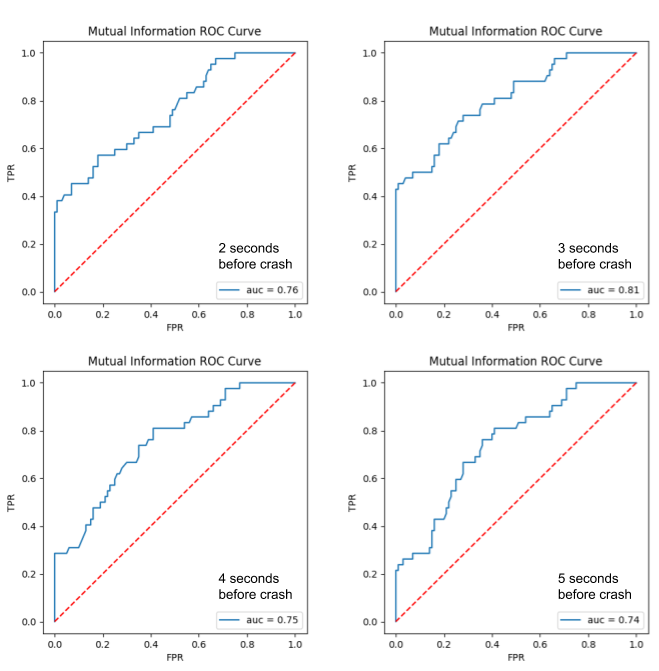}
\caption{The ROC curves 2,3,4,5 seconds before a ``crashed'' or ``not crashed'' event for mutual information.}
\label{figure:rocsecondsbefore}
\end{figure}

This allows us to conclude that
mutual information is 
a promising indicator of incorrect behaviour in real-time, and a time from highest peak to crash of 3 seconds could be sufficient to take an appropriate action. 

\begin{table}
\caption{Distance in frames and seconds from crash.}
  \begin{tabularx}{0.5\textwidth}{XXXXX}
  \hline
  \multicolumn{1}{X}{\textbf{Crash \#}} & \multicolumn{2}{>{\hsize=\dimexpr2\hsize+2\tabcolsep+\arrayrulewidth\relax}X}{\textbf{Distance to first threshold breach}} & \multicolumn{2}{>{\hsize=\dimexpr2\hsize+2\tabcolsep+\arrayrulewidth\relax}X}{\textbf{Distance to defined peak}}\\
  & (frames) & (seconds) & (frames) & (seconds)\\ \hline
1 & 21& 3.5 & 18& 3 \\
2 & 45& 7.5 & 31& 5.17 \\
3 & 42& 7 & 16& 2.7 \\
4 & 39& 6.5 & 7& 1.17 \\
5 & 40& 6.7 & 27& 4.5 \\ \hline
\end{tabularx}
\label{table:timebeforecrash}
\end{table}

\section{Conclusion}
In this paper, we explored the use of uncertainty in end-to-end controllers for self-driving cars and suggested new evaluations for it. We studied 
two separate architectures, one for regression and one for classification, along with the tools to retrieve different types of uncertainty information from them. We tested those architectures in the Udacity self-driving car simulator and found that mutual information, above all other uncertainty measures, is a promising predictor for erroneous behaviour (crashes). All of the above runs in real-time and we thus believe that uncertainty information could play a major role in improving end-to-end controllers and in bringing them up to speed with more traditional and better performing modular controllers.
Planned future work includes varying the speed of the car to determine whether any uncertainty measure is viable at higher speeds, as well as modifying the simulator to include a top down view and the ability to set the car speed directly, and finally exploring a wider range of network architectures. It would also be interesting to evaluate the techniques on real data.

\bibliographystyle{plain}
\bibliography{references}
\end{document}